\documentclass[twoside,lettersize,journal]{IEEEtran}
\usepackage{amsmath,amsfonts}
\usepackage{algorithmic}
\usepackage{algorithm}
\usepackage{array}
\usepackage[caption=false,font=normalsize,labelfont=sf,textfont=sf]{subfig}
\usepackage{textcomp}
\usepackage{stfloats}
\usepackage{url}
\usepackage{verbatim}
\usepackage{graphicx}
\usepackage{multirow}
\usepackage{colortbl}
\usepackage{cite}
\usepackage{adjustbox}
\usepackage{booktabs}
\usepackage{enumitem}
\usepackage{xcolor}

\newcommand{\figref}[1]{Fig. \ref{#1}}
\newcommand{\tabref}[1]{Tab. \ref{#1}}

\newcommand{\secref}[1]{Section \ref{#1}}

\usepackage{hyperref}
\def\BibTeX{{\rm B\kern-.05em{\sc i\kern-.025em b}\kern-.08em
    T\kern-.1667em\lower.7ex\hbox{E}\kern-.125emX}}
\usepackage{balance}

\graphicspath{{./Imgs/}}

\newcommand{\sArt}{state-of-the-art }
\def\eg{\emph{e.g.}}
\def\ie{\emph{i.e.}}  
\def\etal{{\em et al.}}

\begin{document}

\title{Interactive Spatial-Frequency Fusion Mamba for Multi-Modal Image Fusion}
\author{Yixin Zhu,
        Long Lv,
        Pingping~Zhang,
        Xuehu Liu,
        Tongdan Tang,
        Feng~Tian,
        Weibing Sun,
        and~Huchuan~Lu

\thanks{
Yixin~Zhu and Long Lv have equal contributions and are co-first authors.
(Corresponding authors: Pingping Zhang and Tongdan Tang.)

Yixin~Zhu and Pingping~Zhang are with the School of Future Technology, Dalian University of Technology and the Key Laboratory of Data Science and Smart Education (Hainan Normal University), Ministry of Education. (Email: lcabert1@mail.dlut.edu.cn; zhpp@dlut.edu.cn)

Long Lv, Feng Tian and Weibing Sun are with the Affiliated Zhongshan Hospital of Dalian University. (Email: lvlong113@126.com; tianfeng73@163.com; massurm@163.com)

Xuehu Liu is with the School of Computer Science and Artificial Intelligence, Wuhan University of Technology. (Email: liuxuehu@whut.edu.cn)

Tongdan Tang is with the Central Hospital of Dalian University of Technology. (Email: tangtongdan2002@sina.com)

Huchuan~Lu is with the School of Information and Communication Engineering, Dalian University of Technology. (Email: lhchuan@dlut.edu.cn)
}
\thanks{}}

\markboth{IEEE TRANSACTIONS ON IMAGE PROCESSING} 
{Zhu \MakeLowercase{\textit{et al.}}: Interactive Spatial-Frequency Fusion Mamba for Multi-Modal Image Fusion}


\maketitle

\begin{abstract}
Multi-Modal Image Fusion (MMIF) aims to combine images from different modalities to produce fused images, retaining texture details and preserving significant information.
Recently, some MMIF methods incorporate frequency domain information to enhance spatial features.
However, these methods typically rely on simple serial or parallel spatial-frequency fusion without interaction.
In this paper, we propose a novel Interactive Spatial-Frequency Fusion Mamba (ISFM) framework for MMIF.
Specifically,  we begin with a Modality-Specific Extractor (MSE) to extract features from different modalities.
It models long-range dependencies across the image with linear computational complexity.
To effectively leverage frequency information, we then propose a Multi-scale Frequency Fusion (MFF).
It adaptively integrates low-frequency and high-frequency components across multiple scales, enabling robust representations of frequency features.
More importantly, we further propose an Interactive Spatial-Frequency Fusion (ISF).
It incorporates frequency features to guide spatial features across modalities, enhancing complementary representations.
Extensive experiments are conducted on six MMIF datasets.
The experimental results demonstrate that our ISFM can achieve better performances than other state-of-the-art methods.
The source code is available at https://github.com/Namn23/ISFM.
\end{abstract}


\begin{figure}[t]
\centering
  \includegraphics[width=1.0\linewidth]{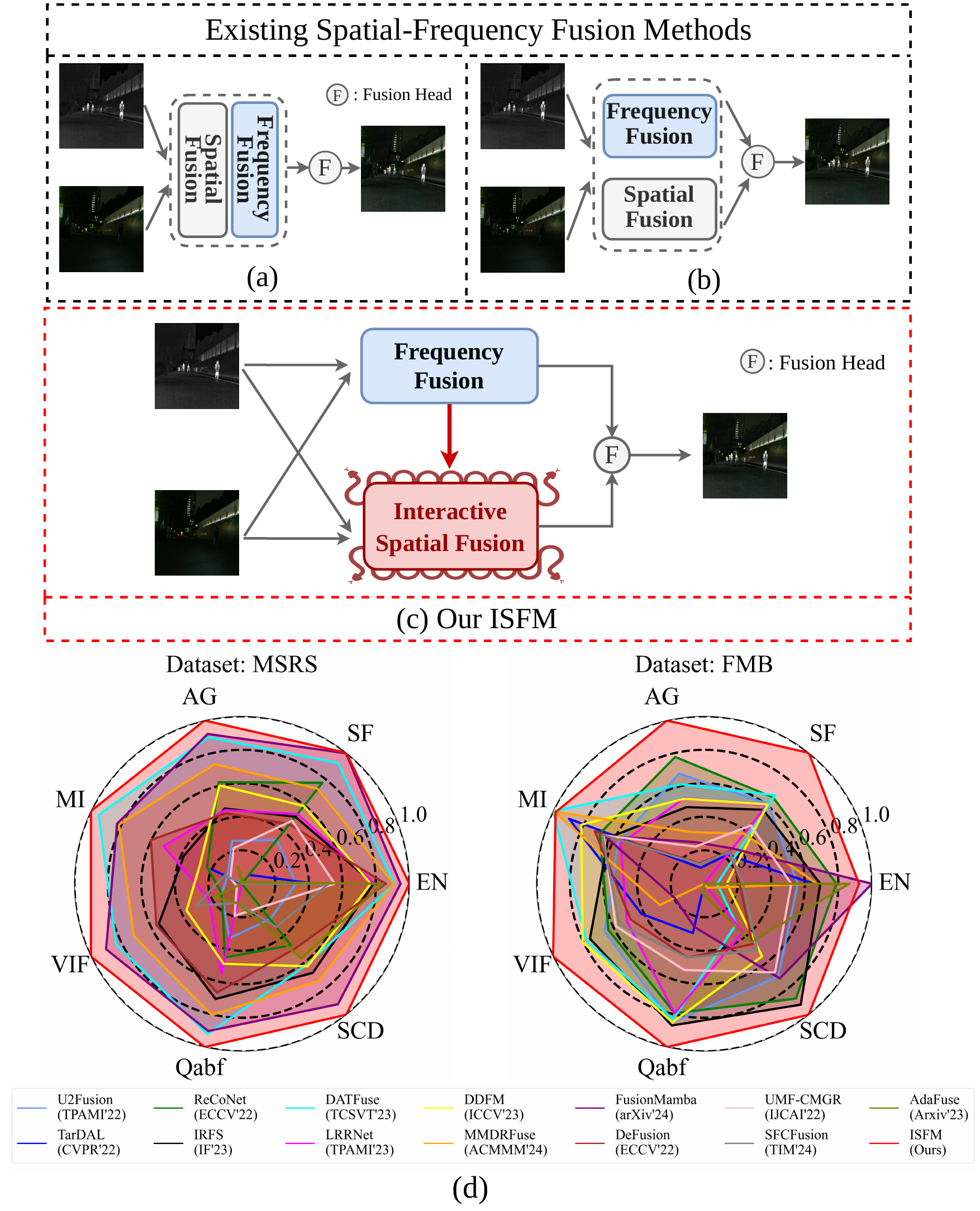}
  \caption{The paradigm and performance comparison of our proposed ISFM and existing MMIF methods. (a) Sequential spatial-frequency fusion methods; (b) Parallel spatial-frequency fusion methods; (c) Our proposed ISFM; (d) Performance comparison on MSRS \cite{tang2022piafusion} and FMB \cite{liu2023multi} in seven metrics.}
  \label{fig:motivation}
\end{figure}
\section{Introduction}
\IEEEPARstart{M}{ulti-Modal} Image Fusion (MMIF) aims to generate more informative and comprehensive images by integrating images from multiple modalities.
Infrared-Visible Image Fusion (IVIF) and Medical Image Fusion (MIF) are two typical MMIF tasks.
Technically, IVIF aims to retain the thermal radiation information from infrared images and fine texture details from visible images.
It overcomes the sensitivity of visible images to various illuminations as well as the sensitivity of infrared images to noises.
MIF can integrate soft tissues from Magnetic Resonance Imaging (MRI) and bone structures from Computed Tomography (CT) modalities.
It provides complementary insights and improves medical diagnostic capabilities.
With these advantages, MMIF can enhance the performance of many downstream tasks, \eg, object detection~\cite{liu2022target}, semantic segmentation~\cite{liu2023multi,kirillov2023segment}.

The primary challenges in MMIF lie in the inherent gaps between modalities and the effective extraction of complementary information.
Recently, some methods~\cite{zheng2024frequency,gu2023adafuse,wu2023dcfusion,hu2024sfdfusion,hou2024infrared,chen2024sfcfusion} start to jointly utilize spatial and frequency information to address these challenges and yield a notable progress.
As illustrated in \figref{fig:motivation}(a), some methods~\cite{zheng2024frequency,wu2023dcfusion} propose to fuse spatial and frequency information sequentially.
Meanwhile, as illustrated in \figref{fig:motivation}(b), some methods~\cite{gu2023adafuse,hu2024sfdfusion,hou2024infrared,chen2024sfcfusion} employ parallel spatial-frequency fusion.
Both of them fail to explore the interaction between the spatial and frequency domains.
They lead to incomplete feature extraction and inefficient exploitation of domain-specific characteristics.
In fact, the spatial domain captures fine-grained texture details while the frequency domain reveals global patterns.
The interactions between two domains ensure a more comprehensive information integration, improving the overall quality of the fused image.

Additionally, existing methods~\cite{liang2022fusion,li2023lrrnet,huang2022reconet,tang2023divfusion,tang2022matr,zhao2023cddfuse,zhao2024equivariant} primarily employ Convolutional Neural Networks (CNNs) or Transformers for feature extraction.
CNNs are inherently limited to model long-range dependencies.
Transformers require significant computational costs.
Recently, Mamba \cite{gu2023mamba} has shown superior performance in vision tasks, due to its ability to model long-range dependencies with linear computational complexity.
Therefore, as illustrated in \figref{fig:motivation}(c), we propose a Mamba-based interactive spatial-frequency fusion framework to fully exploit the complementarity of domain-specific characteristics.

In this paper, we introduce an Interactive Spatial-Frequency Fusion Mamba framework named ISFM for MMIF.
It incorporates frequency information into the spatial fusion process and leverages Mamba to capture long-range dependencies.
More specifically, we begin with a Modality-Specific Extractor (MSE) for modality-specific feature extraction.
Then, we propose a Multi-scale Frequency Fusion (MFF) to effectively leverage frequency information and enable robust representations of frequency features.
The MFF adaptively integrates low-frequency and high-frequency components of different modalities in multiple scales.
To fully explore the complementarity of domain-specific characteristics, we propose an Interactive Spatial-Frequency Fusion (ISF).
Our ISF comprises two parts: a Frequency-Guided Mamba (FGM) and a Frequency-Guided Gate (FGG).
FGM leverages Mamba's ability to enhance complementary representations.
FGG incorporates frequency features to guide the fusion of spatial features across modalities.
By combining MFF and ISF, our ISFM can comprehensively integrate complementary information in the spatial and frequency domains.
Extensive experiments on six MMIF datasets demonstrate that our method can achieve better performances than other state-of-the-art methods.

Our contributions can be summarized as follows:
\begin{itemize}
\item We introduce a novel Interactive Spatial-Frequency Fusion Mamba (ISFM) framework for MMIF. It provides a distinct perspective for spatial-frequency fusion.
\item We propose a Multi-scale Frequency Fusion (MFF) to effectively fuse frequency information across multiple scales. In addition, we propose an Interactive Spatial-Frequency Fusion (ISF) to fully exploit the complementarity of spatial-frequency information.
\item Extensive experiments on IVIF and MIF tasks validate the effectiveness of our method. We also validate our method in helping high-level computer vision tasks.
\end{itemize}
\section{Related Work}
\subsection{Multi-Modal Image Fusion}
MMIF aims to produce informative fused images by integrating the important information from source ones.
Existing MMIF methods can be coarsely categorized into three groups: CNN-based methods\cite{liang2022fusion,li2023lrrnet,huang2022reconet,tang2023divfusion}, Generative Adversarial Network (GAN)-based methods \cite{liu2022target,ma2019fusiongan,ma2020infrared}, and Transformer-based methods\cite{ma2022swinfusion,tang2022matr,zhao2023cddfuse,zhao2024equivariant}.
CNN-based methods utilize CNNs to achieve image fusion. They are usually optimized through elaborated loss functions and network architectures.
For example, Ma \etal \cite{ma2021stdfusionnet} develop a mask loss to enable a selective integration of target and background regions.
Zhang \etal \cite{zhang2020rethinking} design a uniform form of loss functions to generate the final fused image.
However, due to the local receptive field, CNN-based methods struggle to model long-range dependencies.
GAN-based methods leverage GAN\cite{goodfellow2014generative} to ensure that the fused images are distributionally similar to the input images.
Ma \etal \cite{ma2019fusiongan} first apply GAN to the field of image fusion.
They utilize the discriminator to guide the generator in producing texture-rich fused images.
Additionally, Ma \etal \cite{ma2020ddcgan} introduce a dual discriminator conditional GAN to achieve a balanced fusion, where both infrared and visible images contribute to the adversarial process.
However, GAN-based methods have the model instability and collapse problem.
Furthermore, researchers have explored Transformer-based methods.
For example, Zhao \etal \cite{zhao2023cddfuse} merge CNNs with Transformers to decompose and reconstruct image features. %
Tang \etal \cite{ma2022swinfusion} utilize SwinTransformer \cite{liu2021swin} to overcome the shortcomings of CNN-based fusion.
Recent studies \cite{zhang2025omnifuse,yi2025artificial,cao2025efficient} further advance the robustness and semantic consistency, providing new insights for improving MMIF.
However, they face significant computational challenges.
Different from these methods, we propose a Mamba-based spatial-frequency fusion framework for MMIF.
It incorporates frequency information into the spatial fusion process and leverages Mamba to capture long-range dependencies.
\subsection{Frequency Domain Representations in Image Fusion}
Frequency domain representations can highlight the underlying global patterns and structural characteristics of an image~\cite{tancik2020fourier}.
Therefore, several studies~\cite{wang2024efficient,bhat2017feature,balasubramanian2008new,hu2024sfdfusion,hou2024infrared,chen2024sfcfusion,wu2023dcfusion,li2013region} utilize the frequency domain information for MMIF.
Early works simply incorporate frequency information during the fusion process.
For example, Balasubramanian \etal \cite{balasubramanian2008new} use wavelet packet transform and employ the fusion of high-frequency sub-bands.
Bhat \etal \cite{bhat2017feature} apply principle component analysis to frequency domain transform for medical image fusion.
Wang \etal \cite{wang2024efficient} design a frequency domain auto-encoder to achieve image fusion.
Recently, some researchers start to integrate spatial and frequency domain information for MMIF.
For example, Zheng \etal \cite{zheng2024frequency} propose to fuse spatial and frequency domain features sequentially.
Gu \etal \cite{gu2023adafuse} propose a cross attention module to fuse cross-domain information in parallel.
Hu \etal \cite{hu2024sfdfusion} construct a parallel spatial and frequency fusion network to guide the fusion process.
Chen \etal \cite{chen2024sfcfusion} decompose images in the frequency domain and employ a parallel fusion process.
Wu \etal \cite{wu2023dcfusion} design a cascaded spatial and frequency enhancement module for image fusion.
However,  these methods typically integrate information of spatial and frequency domains without interaction.
Different from them, we propose an Interactive Spatial-Frequency Fusion (ISF) to incorporate frequency information into the spatial fusion process.
It leverages Mamba to extract complementary features from spatial and frequency domains.
\subsection{Mamba in Computer Vision}
Recently, Mamba \cite{gu2023mamba} introduces a selection mechanism for modeling long sequences with linear computational complexity.
It has received significant attention in the computer vision community.
For instance, Vision Mamba \cite{zhu2024vision} introduces a bidirectional Mamba approach that enables global perception.
Liu \etal \cite{liu2024vmamba} further propose VMamba to achieve more spatial connections through a four-directional scanning strategy.
Guo \etal \cite{guo2025mambair} design MambaIR, which introduces a residual state space block for image restoration.
He \etal \cite{he2025pan} propose Pan-Mamba to tackle the challenge of pan-sharpening.
For MMIF, Xie \etal \cite{xie2024fusionmamba} propose a dynamic feature enhancement with Mamba to achieve MMIF.
Zhu \etal \cite{zhu2024mamba} propose a dual-branch network consisting of Transformer and Mamba.
Cao \etal \cite{cao2024shuffle} propose a random scanning strategy to tackle the fixed sequence scanning.
However, these methods are limited to spatial domain fusion.
In this paper, we propose a multi-scale frequency fusion to effectively leverage frequency information.
In addition, we propose a Mamba-based interactive spatial-frequency fusion to integrate complementary information in the spatial and frequency domains.
\begin{figure*}[t]
\centering
\includegraphics[width=0.9\linewidth]{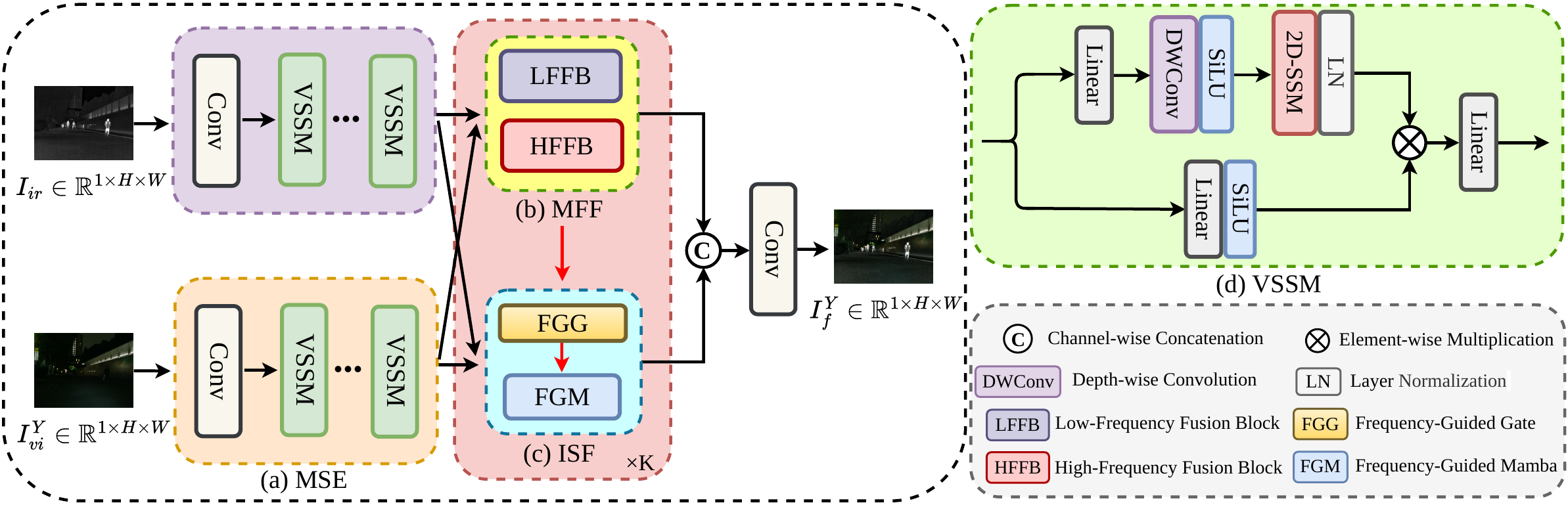}
\caption{Overview of our proposed ISFM framework. (a) Modality-Specific Extractor (MSE) extracts modality-specific features; (b) Multi-scale Frequency Fusion (MFF) employs frequency domain fusion in different scales; (c) Interactive Spatial-Frequency Fusion (ISF) incorporates frequency information into the spatial fusion; (d) Structure of Vision State-Space Module (VSSM).}
\label{fig:overall}
\end{figure*}
\section{Methodology}
\label{sec:method}
\subsection{Overall Architecture}
\label{sec:overall}
As shown in \figref{fig:overall}, our framework comprises two branches to implement spatial-frequency fusion.
Note that, it is a general MMIF framework.
Without loss of generality, we take the IVIF task as an example to showcase its workflow.
Given a pair of infrared image \( I_{ir} \in \mathbb{R}^{1 \times H \times W} \) and visible image \( I_{vi} \in \mathbb{R}^{3 \times H \times W} \), the RGB-to-YCbCr color conversion \cite{tang2023divfusion} is utilized to solve the channel mismatching problem.
We first extract the Y channel of the visible image, denoted as \( I_{vi}^Y \in \mathbb{R}^{1 \times H \times W} \).
Here, \(H\) and \(W\) represent the height and width of input images, respectively.
Then, we employ the Mamba-based Modality-Specific Extractor (MSE) to extract modality-specific features.
Afterwards, we process the features by the Multi-scale Frequency Fusion (MFF) and Interactive Spatial-Frequency Fusion (ISF).
MFF is proposed to perform the adaptive fusion of low-frequency and high-frequency features.
ISF incorporates frequency features to guide the spatial fusion process, facilitating the comprehensive extraction of complementary features.
Finally, features from spatial and frequency domains are concatenated along the channel dimension.
We employ two convolutional layers to produce the Y channel \(I_{f}^Y \in \mathbb{R}^{1 \times H \times W}\) of the final fused image.
\subsection{Modality-Specific Extractor}
\label{sec:MSE}
To extract modality-specific features from different modalities, we employ a Mamba-based Modality-Specific Extractor (MSE).
More specifically, given the inputs \(I_{ir}\) and \(I_{vi}^Y\), we first apply two \(3\times3\) convolutional layers to generate the shallow features \( \ F_{ir}^S \in \mathbb{R}^{C \times H \times W}\) and \( F_{vi}^S \in \mathbb{R}^{C \times H \times W}\).
Here, \(C\) represents the channel dimension of features. The process is expressed as follows:
\begin{equation}
\begin{aligned}
F_{ir}^S &=  SiLU(Conv_{3\times3}(I_{ir})), \\
F_{vi}^S &=  SiLU(Conv_{3\times3}(I_{vi}^Y)),
\label{eq:shallow}
\end{aligned}
\end{equation}
where \(SiLU\) is the SiLU activation function \cite{shazeer2020glu}.

Then, we employ the Vision State-Space Module (VSSM) \cite{liu2024vmamba} to extract high-level features.
It effectively captures long-range dependencies of image features.
Note that the wights of two modalities are not shared.
By adopting CNN and Mamba, MSE produces modality-specific features \( F_{ir}^M, F_{vi}^M \in \mathbb{R}^{C \times H \times W} \).
This process is expressed by:
\begin{equation}
\begin{aligned}
F_{ir}^M &=  \Psi_{ir}(  F_{ir}^S),\\
F_{vi}^M &= \Psi_{vi}(F_{vi}^S),
\label{eq:ms}
\end{aligned}
\end{equation}
where \(\Psi\) represents the VSSM blocks.
\subsection{Multi-Scale Frequency Fusion}
\label{sec:FFB}
\begin{figure}[t]
\centering
\includegraphics[width=0.9\linewidth]{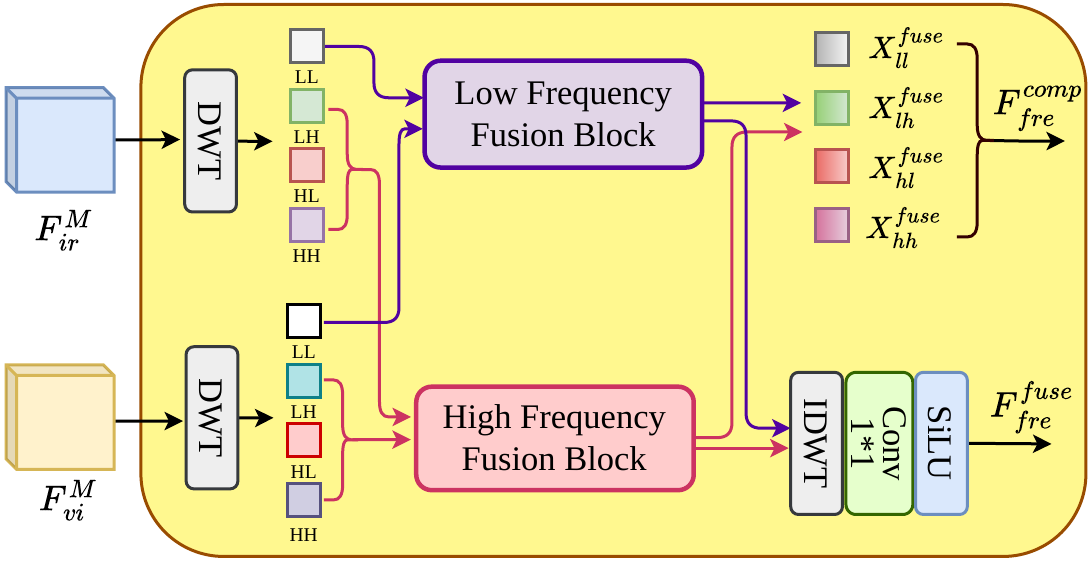}
\caption{Illustration of the proposed MFF.}
\label{fig:mff}
\end{figure}

Frequency domain representations capture global characteristics and reveal the underlying patterns of an image~\cite{tancik2020fourier}.
This allows for a more systematic decomposition of image details, enabling more precise feature extraction and analysis.
Therefore, we propose a Multi-scale Frequency Fusion (MFF) to effectively integrate the frequency information.
As illustrated in \figref{fig:mff}, we apply the Discrete Wavelet Transform (DWT) \cite{mallat1989theory} to obtain low-frequency components $LL$ and high-frequency components $LH$, $HL$ and $HH$.
We then propose a Low-Frequency Fusion Block (LFFB) and a High-Frequency Fusion Block (HFFB), which implement the adaptive fusion of low-frequency and high-frequency components, respectively.
\subsubsection{Low-Frequency Fusion Block}
\label{LFFB}
\begin{figure}[t]
\centering
\includegraphics[width=0.9\linewidth]{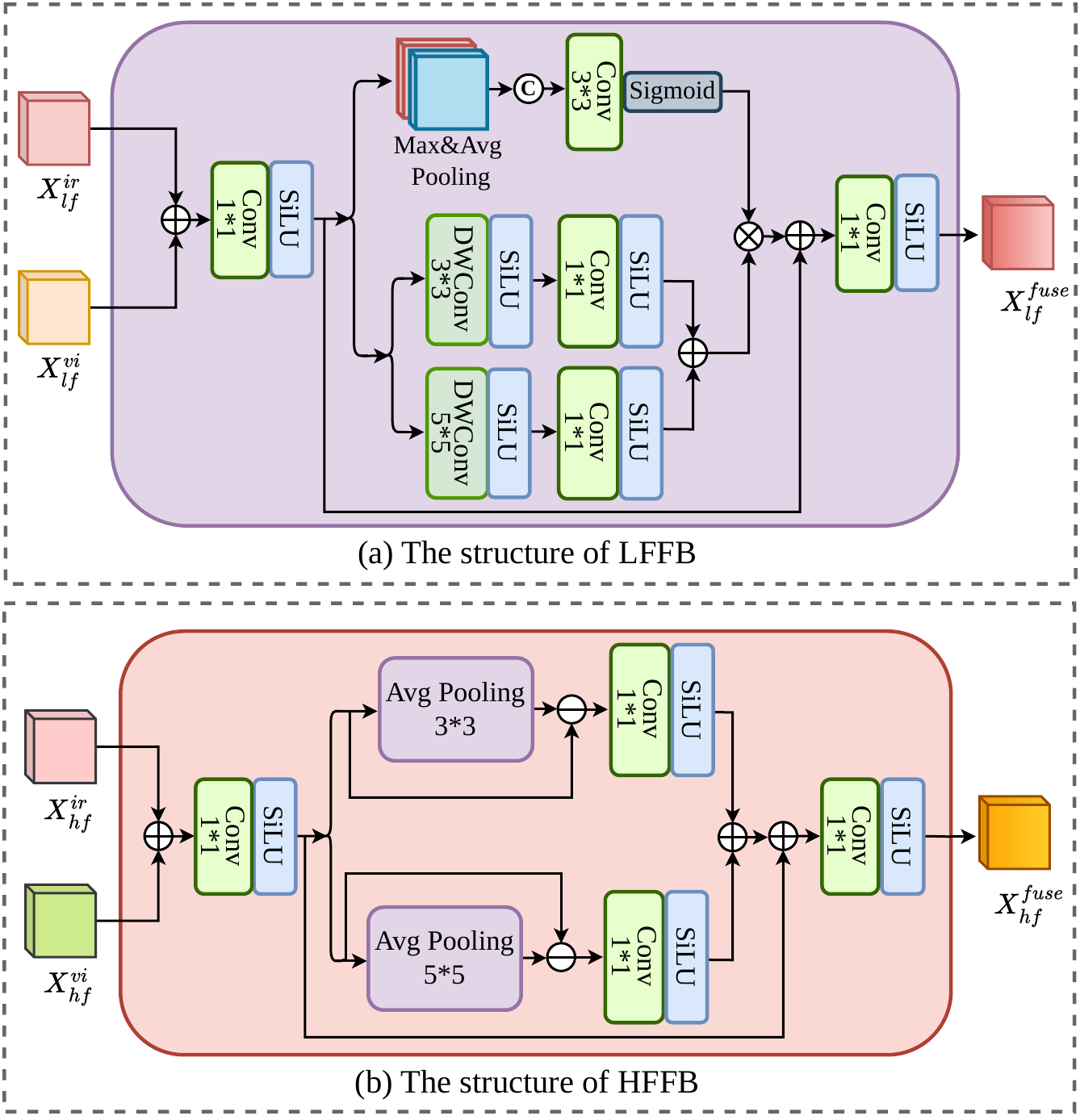}
\caption{Illustration of the proposed LFFB and HFFB.}
\label{fig:lhffb}
\end{figure}

As illustrated in \figref{fig:lhffb}(a), we first perform an element-wise addition on the low-frequency components.
We then employ an \({1\times1}\) convolutional layer to generate \(X_{lf}\), which is expressed as follows:
\begin{equation}
\begin{aligned}
X_{lf}^{ir}  &= DWT(F_{ir}^M),X_{lf}^{vi} = DWT(F_{vi}^M),\\
X_{lf} &= SiLU(Conv_{1\times1}(X_{lf}^{ir} + X_{lf}^{vi})).
\label{eq:lf}
\end{aligned}
\end{equation}

Subsequently, we process the combined components through two streams.
The first stream applies the max pooling and average pooling layers to adaptively capture the global information of \(X_{lf}\).
We then conduct a \({3\times3}\) convolutional layer and a sigmoid layer to compute spatial attention weights \(\sigma_{s}\).
This process is expressed by:
\begin{equation}
\begin{aligned}
\sigma_{s} = Sigmoid(Conv_{3\times3}([\mathbf{\mathcal{M}}(X_{lf});\mathbf{\mathcal{A}}(X_{lf})])),
\label{eq:mapool}
\end{aligned}
\end{equation}
where \(\mathbf{\mathcal{M}(\cdot)}\) and \(\mathbf{\mathcal{A}(\cdot)}\) represents the max pooling and average pooling layers, respectively.

The second stream utilizes the depth-wise convolutional layers with different kernel sizes to process \(X_{lf}\).
We then employ an \({1\times1}\) convolutional layer to generate features at different scales.
The process is expressed by:
\begin{equation}
\begin{aligned}
U' &= SiLU(Conv_{1\times1}(DWConv_{3\times3}(X_{lf}))),\\
U'' &= SiLU(Conv_{1\times1}(DWConv_{5\times5}(X_{lf}))),
\label{eq:dwconv}
\end{aligned}
\end{equation}
where \(DWConv_{3\times3}(\cdot)\) and \(DWConv_{5\times5}(\cdot)\) represent a \(3\times3\) and \(5\times5\) depth-wise convolutional layer, respectively.

Given different scales of contextual information, we utilize attention weights \(\sigma_{s}\) to extract significant information through an element-wise multiplication.
We then apply a residual connection to \(X_{lf}\).
Finally, we pass the aggregated features through an \({1\times1}\) convolutional layer to obtain the fused low-frequency features \(X_{lf}^{fuse}\).
The process is expressed by:
\begin{equation}
\begin{aligned}
U &= (U' + U'') \otimes \sigma_{s} + X_{lf},\\
X_{lf}^{fuse} &= SiLU(Conv_{1\times1}(U)),
\label{eq:lffuse}
\end{aligned}
\end{equation}
where \(\otimes\) denotes the element-wise multiplication.
\subsubsection{High-Frequency Fusion Block}
\label{HFFB}
As illustrated in \figref{fig:lhffb}(b), we employ the element-wise addition on the high-frequency components.
We then apply an \({1\times1}\) convolutional layer to generate \(X_{hf}\), which is expressed by:
\begin{equation}
\begin{aligned}
X_{hf}^{ir}  &= DWT(F_{ir}^M),X_{hf}^{vi} = DWT(F_{vi}^M),\\
X_{hf} &= SiLU(Conv_{1\times1}(X_{hf}^{ir} + X_{hf}^{vi})).
\label{eq:hf}
\end{aligned}
\end{equation}

To enhance the edge details of \(X_{hf}\), we apply two average pooling layers with different sizes.
The process is formulated as follows:
\begin{equation}
\begin{aligned}
X_{hf}' &= \mathcal{A}_{3\times3}(X_{hf}),
X_{hf}'' &=\mathcal{A}_{5\times5}(X_{hf}),
\label{eq:avg}
\end{aligned}
\end{equation}
where \(\mathcal{A}_{3\times3}(\cdot)\) and \(\mathcal{A}_{5\times5}(\cdot)\) denote a \(3\times3\) average pooling layer and a \(5\times5\) average pooling layer, respectively.

Subsequently, we subtract the pooled results from the original high-frequency features.
This operation accentuates the finer, more salient edge details by emphasizing the discrepancies between the original high-frequency information and the smoothed versions \cite{yan2023transy}.
We then adapt an \({1\times1}\) convolutional layer to generate \(S_1\) and \(S_2\).
The process is expressed by:
\begin{equation}
\begin{aligned}
S_1 &= SiLU(Conv_{1\times1}(X_{hf} - X_{hf}')),\\
S_2 &= SiLU(Conv_{1\times1}(X_{hf} - X_{hf}'')).
\label{eq:hf_s}
\end{aligned}
\end{equation}

In high-frequency features, noises typically appear as random patterns, while edge details are structured.
By subtracting the average-pooled features from the original high-frequency features, edge details are enhanced and noises are suppressed.
Finally, we sum up the features and apply a residual connection to \(X_{hf}\).
We then pass them through an \({1\times1}\) convolutional layer to obtain the fused high-frequency features \(X_{hf}^{fuse}\).
The process is formulated as follows:
\begin{equation}
\begin{aligned}
S &= S_1 + S_2,\\
X_{hf}^{fuse} &= SiLU(Conv_{1\times1}(S+X_{hf})).
\label{eq:hf_fuse}
\end{aligned}
\end{equation}

Through the combination of LFFB and HFFB, the fused frequency features are passed through two operations.
First, they are utilized as frequency-domain guidances for the ISF, denoted as \(F_{fre}^{comp}\).
It enables an interactive fusion to fully exploit the complementary information in spatial and frequency domains, which will be detailed in \secref{sec:FISF}.
Second, the fused frequency features are converted into the spatial domain via inverse DWT followed by a convolutional layer, denoted as \(F_{fre}^{fuse}\).
It will be combined with the fused spatial features to generate the final fused image.
\begin{figure*}[t]
\centering
\includegraphics[width=0.9\linewidth]{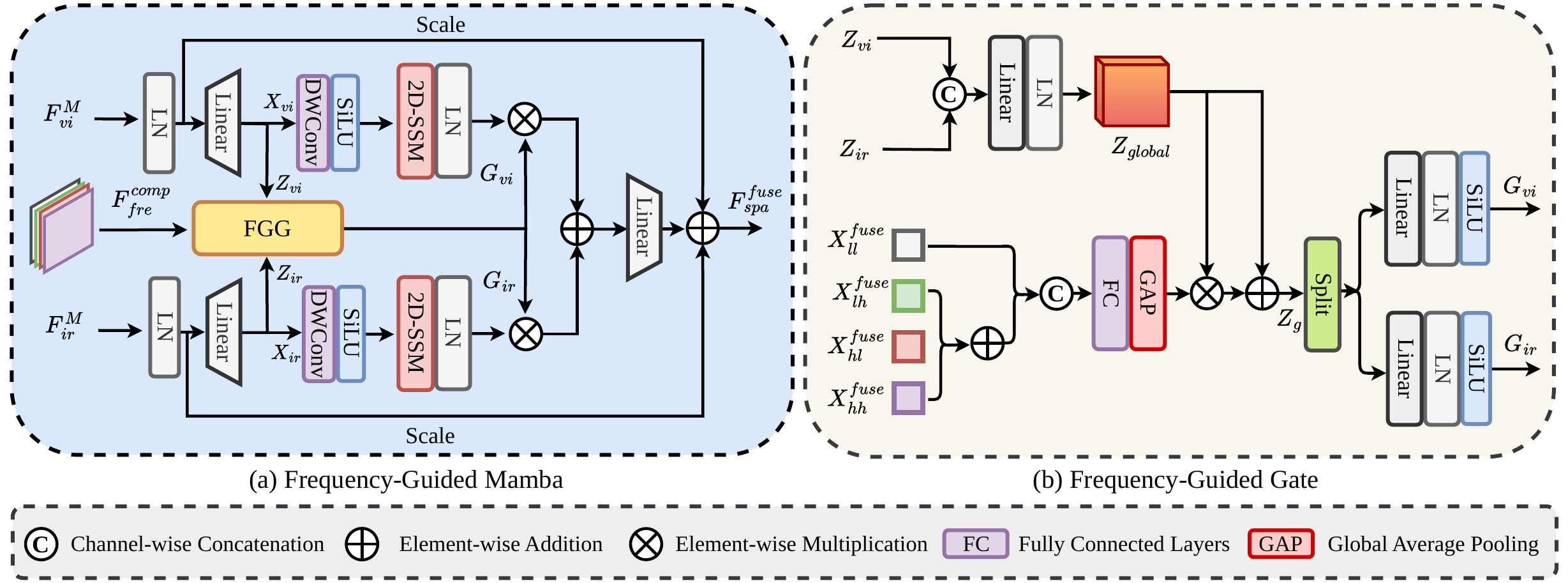}
\caption{The architecture of the proposed ISF. (a) Frequency-Guided Mamba (FGM) leverages Mamba to enhance complementary representations; (b) Frequency-Guided Gate (FGG) incorporates frequency features to guide spatial features across modalities.}
\label{fig:FISF}
\end{figure*}
\subsection{Interactive Spatial-Frequency Fusion}
\label{sec:FISF}
The Interactive Spatial-Frequency Fusion (ISF) aims to take advantage of the frequency features to achieve comprehensive extraction of complementary features across modalities.
It is implemented by a Frequency-Guided Mamba (FGM) and a Frequency-Guided Gate (FGG).
\subsubsection{Frequency-Guided Mamba}
\label{sec:FGCM}
As illustrated in \figref{fig:FISF}(a), the input features \(F_{ir}^M\) and \(F_{vi}^M\) are first passed through a normalization layer.
Taking the infrared stream as an example, the features are then processed by a linear layer and split along the channel dimension.
It generates two distinct feature sequences, \ie, \(X_{ir} \in \mathbb{R}^{ \eta C \times H \times W} \) and \(Z_{ir} \in \mathbb{R}^{\eta C \times H \times W}\).
Here, \(\eta\) is the channel expansion factor.
This procedure can be expressed by:
\begin{equation}
\begin{aligned}
X_{ir} &= Linear(LN(F_{ir}^M)),Z_{ir} = Linear(LN(F_{ir}^M)),\\
X_{vi} &= Linear(LN(F_{vi}^M)),Z_{vi} = Linear(LN(F_{vi}^M)),
\label{eq:linear_z}
\end{aligned}
\end{equation}
where \(Linear(\cdot)\) and \(LN(\cdot)\) represent the linear layer and the layer normalization~\cite{xiong2020layer}, respectively.

Afterwards, \(X_{ir}\) is processed through a depth-wise convolutional layer.
Then, it is fed into the 2D-SSM block and a normalization layer to produce the hidden state features \(H_{ir}\).
The process is expressed as follows:
\begin{equation}
\begin{aligned}
H_{ir} &= LN(SSM(SiLU(DWConv_{3\times3}(X_{ir})))),\\
H_{vi} &= LN(SSM(SiLU(DWConv_{3\times3}(X_{vi})))).
\label{eq:ssm}
\end{aligned}
\end{equation}

Meanwhile, the initial feature \(Z_{ir}\) interacts with the frequency features through FGG.
It generates the gating features to guide more robust spatial fusion, which will be detailed in \secref{sec:FGGM}.
The process is expressed as follows:
\begin{equation}
\begin{aligned}
G_{ir}, G_{vi} = \mathbf{\mathcal{G}}(Z_{ir}, Z_{vi}, F_{freq}^{comp}),
\label{eq:ssm_g}
\end{aligned}
\end{equation}
where \(\mathbf{\mathcal{G}(\cdot)}\) denotes FGG.

Subsequently, we utilize \(G_{ir}\) to modulate \(H_{ir}\) to achieve the interactive integration of spatial and frequency information.
Finally, we apply the tunable scale factors to the input features.
They are used as residual connections to generate the fused spatial feature \(F_{spa}^{fuse}\).
This process is formulated by:
\begin{equation}
\begin{aligned}
F_h &= H_{ir} \otimes G_{ir} + H_{vi} \otimes G_{vi},\\
F_{spa}^{fuse} &= Linear(F_h) + s_1 \cdot F_{ir}^M + s_2 \cdot F_{vi}^M,
\label{eq:final_fgcm}
\end{aligned}
\end{equation}
where \(s_1\) and \(s_2\) represent the tunable scale factors.
\subsubsection{Frequency-Guided Gate}
\label{sec:FGGM}
To take advantage of frequency complementarities, we propose a Frequency-Guided Gate (FGG).
It enhances Mamba's ability to extract complementary information across modalities.
As illustrated in \figref{fig:FISF}(b), we first conduct the channel-wise concatenation on the features \(Z_{ir}\) and \(Z_{vi}\).
We then adopt a linear layer to generate \(Z_{global}\), which is expressed by:
\begin{equation}
\begin{aligned}
Z_{global} = LN(Linear[Z_{ir}; Z_{vi}]).
\label{eq:z_global}
\end{aligned}
\end{equation}

Given the frequency components, we first sum up the high-frequency components and concatenate them with the low-frequency components.
We then use fully connected layers and a global average pooling layer to aggregate channel information.
It encodes global statistics of the frequency components, which is expressed by:
\begin{equation}
\begin{aligned}
F_f &= [X_{ll}^{fuse};X_{lh}^{fuse}+X_{hl}^{fuse}+X_{hh}^{fuse}],\\
Z_f &= GAP(FC(F_f)),
\label{eq:z_f}
\end{aligned}
\end{equation}
where \(FC(\cdot )\) represents the fully connected layers and \(GAP(\cdot)\) denotes the global average pooling layer.

The encoded features are then used to guide the selective fusion of spatial features, which is expressed by:
\begin{equation}
\begin{aligned}
Z_{g} = Z_f \otimes Z_{global} + Z_{global}.
\label{eq:z_g}
\end{aligned}
\end{equation}

Afterwards, we split \(Z_{g}\) along the channel dimension.
Each part is processed through a linear layer, followed by a normalization layer and a SiLU activation function.
This enables more precise and adaptive gating while enhancing complementary features.
The process is expressed by
\begin{equation}
\begin{aligned}
Z_{g}', Z_{g}'' &= Split(Z_{g}),\\
G_{ir} &= SiLU(LN(Linear(Z_{g}'))),\\
G_{vi} &= SiLU(LN(Linear(Z_{g}''))).
\label{eq:gate}
\end{aligned}
\end{equation}
Here, \(G_{ir}\) and \(G_{vi}\) are the generated gating weights, which are derived from different modalities and frequency components.
\begin{figure*}[htbp]
\centering
  \includegraphics[width=1.0\linewidth]{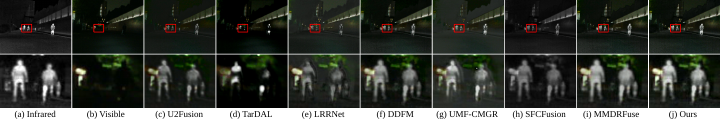}
  \caption{Comparison of fusion results on the MSRS dataset. The second row shows the enlarged regions in the first row.}
  \label{fig:visual_msrs}
\end{figure*}
\begin{figure*}[htbp]
\centering
  \includegraphics[width=1.0\linewidth]{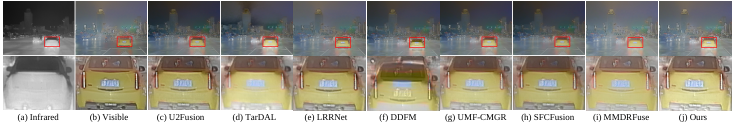}
  \caption{Comparison of fusion results on the FMB dataset. The second row shows the enlarged regions in the first row.}
  \label{fig:visual_fmb}
\end{figure*}
\begin{figure*}[htbp]
\centering
  \includegraphics[width=1.0\linewidth]{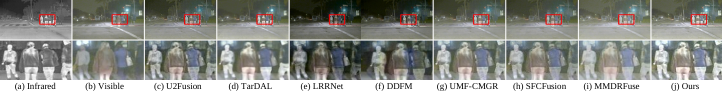}
  \caption{Comparison of fusion results on the RoadScene dataset. The second row shows the enlarged regions in the first row.}
  \label{fig:visual_road}
\end{figure*}
\subsection{Loss Functions}
\label{sec:loss}
The loss function plays a crucial role in balancing the preserving information from different image modalities.
In this paper, the training process is optimized by a combination of content loss \(\mathcal{L}_{cont}\), pixel intensity loss \(\mathcal{L}_{int}\), gradient loss \(\mathcal{L}_{grad}\) and structural similarity loss \(\mathcal{L}_{ssim}\).
The total loss is represented as follows:
\begin{equation}
\begin{aligned}
\mathcal{L}_{total} = \mathcal{L}_{cont} + \alpha \mathcal{L}_{int} + \lambda \mathcal{L}_{grad}+ \gamma \mathcal{L}_{ssim},
\label{eq:total_loss}
\end{aligned}
\end{equation}
where \(\alpha\), \(\lambda\) and \(\gamma\) are the hyper-parameters that control the trade-off of each sub-loss term.

We compute \(\mathcal{L}_{cont}\) as the sum of the differences between the fused image \(I_{f}^Y\) and the source images \(I_{ir}\) and \(I_{vi}^Y \):
\begin{equation}
\begin{aligned}
\mathcal{L}_{cont} = \frac{1}{HW}(\|I_{f}^Y - I_{ir} \|_1 + \|I_{f}^Y - I_{vi}^Y \|_1),
\label{eq:loss_cont}
\end{aligned}
\end{equation}
where $H$ and $W$ are the height and width of an image, respectively.
\(\|\cdot\|_1\) stands for the \(l_1\)-norm.

We adapt \(\mathcal{L}_{int}\) and \(\mathcal{L}_{grad}\) to maximize the preservation of salient targets and detailed texture information from source modalities in the fused image.
It can be expressed by:
\begin{equation}
\begin{aligned}
\mathcal{L}_{int} &= \frac{1}{HW}(\|I_{f}^Y - max(I_{ir}, I_{vi}^Y) \|_1),\\
\mathcal{L}_{grad} &= \frac{1}{HW}(\| | \nabla I_{f}^Y | - max(| \nabla I_{ir} |, | \nabla I_{vi}^Y |)),
\label{eq:loss_grad}
\end{aligned}
\end{equation}
where \(|\cdot|\) and \(\nabla\) denote the absolute value operation and the Sobel gradient operator, respectively.

To comprehensively evaluate the structured quality of the fused image, we employ \(\mathcal{L}_{ssim}\), which is expressed by:
\begin{equation}
\begin{aligned}
\mathcal{L}_{ssim} = \frac{1-SSIM(I_{f}^Y,I_{ir})}{2} + \frac{1-SSIM(I_{f}^Y,I_{vi}^Y)}{2},
\label{eq:loss_ssim}
\end{aligned}
\end{equation}
where \(SSIM(\cdot)\) is the structural similarity index metric \cite{wang2004image}, which measures the structural similarity of two images.
\section{Experiments}
\label{sec:expers}
\begin{figure*}
\centering
  \includegraphics[width=1.0\linewidth]{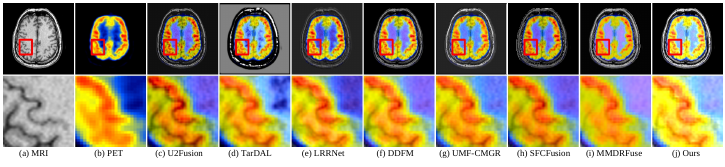}
  \caption{Comparison of fusion results on the MRI-PET Harvard Medical dataset. The second row shows the enlarged regions in the first row.}
  \label{fig:visual_mripet}
\end{figure*}
\begin{figure*}
\centering
  \includegraphics[width=1.0\linewidth]{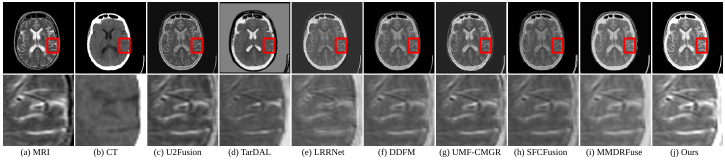}
  \caption{Comparison of fusion results on the MRI-CT Harvard Medical dataset. The second row shows the enlarged regions in the first row.}
  \label{fig:visual_mrict}
\end{figure*}
\begin{figure*}
\centering
  \includegraphics[width=1.0\linewidth]{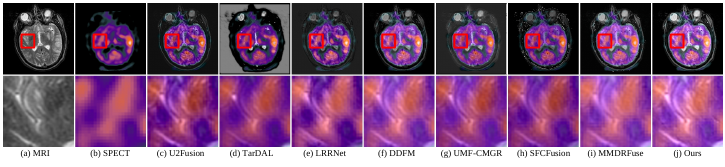}
  \caption{Comparison of fusion results on the MRI-SPECT Harvard Medical dataset. The second row shows the enlarged regions in the first row.}
  \label{fig:visual_mrispect}
\end{figure*}
\begin{table*}[htbp]
\renewcommand{\arraystretch}{0.91}
\centering
\caption{Quantitative results of the IVIF task. The best performance is marked in \textbf{bold} and the second-best performance is \underline{underlined}.}
\begin{adjustbox}{width=0.8\textwidth}
\begin{tabular}{@{}c|l|ccccccccc@{}}
\toprule
\textbf{Dataset} & \textbf{Method} & \textbf{EN(↑)} & \textbf{SF(↑)} & \textbf{AG(↑)} & \textbf{VIF(↑)} &\textbf{MI(↑)} &\(\mathbf{Q^{AB/F}}\)(↑) & \textbf{SCD(↑)} & \textbf{Avg.R(↓)} \\ \midrule
\multirow{7}{*}
& ReCoNet \cite{huang2022reconet}   &  4.23 	&  9.98 	&  2.99 	&  0.49 	&  1.58 	&  0.40 	&  1.26 	& 10.71   \\
& DeFusion \cite{liang2022fusion}      & 6.35 	& 7.98 	& 2.60 	& 0.75 	& 2.16 	& 0.51 	& 1.29 	& 8.71 \\
& U2Fusion \cite{xu2020u2fusion}   & 5.03	& 7.22	& 2.25		& 0.49	& 1.37  & 0.34 	& 1.06 & 13.14 \\
& TarDAL \cite{liu2022target} &  5.32 	& 5.54 	& 1.71 	& 0.41 	& 1.52 	& 0.17 	& 0.79 	& 14.71 \\
& IRFS \cite{wang2023interactively}   & 6.17 	& 8.34 	& 2.66 	& 0.74 	& 1.82 	& 0.53 	& 1.47 	& 8.29 \\
& AdaFuse \cite{gu2023adafuse}      & 6.31	 & 5.19	 & 1.93	 & 0.52	 & 1.22	 & 0.23	 & 1.36	 & 12.57 \\
{\textbf{MSRS}} & DATFuse \cite{tang2023datfuse}   & 6.48 	& 10.93 	& 3.56 	& 0.91 	&\underline{2.70} 	& 0.64 	& 1.41 	& 4.57 \\
& LRRNet \cite{li2023lrrnet}     & 6.19 	& 8.47 	& 2.64 	& 0.54 	& 2.03 	& 0.45 	& 0.79 	& 10.00 \\
& DDFM \cite{zhao2023ddfm}  & 6.13 	& 8.86 	& 2.95 	& 0.63 	& 1.65 	& 0.42 	& 1.42 	& 9.00 \\
& FusionMamba \cite{xie2024fusionmamba}      & 6.57	 & \underline{11.40}	& 3.60	& \underline{0.95}	& 2.51	& 0.63	& \underline{1.71}	& \underline{3.00} \\
& SFCFusion \cite{chen2024sfcfusion}     &5.51	&5.11	&1.72		&0.59  &1.40	 &0.23	&1.12   &13.29 \\
& MMDRFuse \cite{deng2024mmdrfuse}     & 6.47	& 9.69	& 3.22	& 0.84	& 2.49	& 0.58	& 1.53	& 5.57 \\
& UMF-CMGR \cite{wang2022unsupervised}    & 5.60	& 8.15	& 2.16	& 0.43	& 1.34	& 0.27	& 0.97	& 13.29 \\
&FISCNet\cite{zheng2024frequency}	&6.67	&11.36	&3.69	&0.92  &2.44	 &\underline{0.65}&1.55	 &3.43 \\
&RPFNet\cite{guan2025residual}	 &\textbf{7.26}	&11.39	&\underline{3.75}	&0.83 &1.75		&0.59	&1.61 & 4.14 \\
& ISFM(Ours)    & \underline{6.70} & \textbf{11.42} & \textbf{3.77} & \textbf{1.01} & \textbf{2.78} & \textbf{0.68} & \textbf{1.79} & \textbf{1.14} \\ \midrule
\multirow{7}{*}
& ReCoNet \cite{huang2022reconet}  & 6.69 	& 10.70 	& 3.62	& 0.65 	& 2.25 	& 0.55 	& 1.60 	& 6.00  \\
& DeFusion \cite{liang2022fusion}     & 6.39 	& 7.13 	& 2.22 	& 0.59 	& 2.29 	& 0.37 	& 1.33 	& 11.43 \\
& U2Fusion \cite{xu2020u2fusion}   &6.57	&10.27	&3.40		&0.66 &2.10	&0.58	&1.49 & 7.29 \\
& TarDAL \cite{liu2022target}   & 6.62 	& 6.91 	& 2.16 	& 0.56 	& 2.50 	& 0.29 	& 1.03 	& 12.29  \\
& IRFS \cite{wang2023interactively}    & 6.64 	& 9.88 	& 2.96 	& 0.70 	& 2.18 	& 0.59 	& \underline{1.63} 	& 6.86 \\
& AdaFuse \cite{gu2023adafuse}      & 6.73	& 4.98	& 1.95	& 0.39	& 1.40	& 0.16	& 1.32	& 13.57 \\
{\textbf{FMB}}& DATFuse \cite{tang2023datfuse}  & 6.32 	& 10.85 	& 3.26 	& 0.71 	& \underline{2.93} 	& 0.57 	& 1.23 	& 7.43 \\
& LRRNet \cite{li2023lrrnet}  & 6.28 	& 10.18 	& 3.08 	& 0.60 	& 2.08 	& 0.55 	& 1.27 	& 10.86 \\
& DDFM \cite{zhao2023ddfm}   & 6.35	& 10.23	 & 3.07	& 0.72	& 2.40	& 0.58	& 1.39	& 7.29 \\
& FusionMamba \cite{xie2024fusionmamba}    & \underline{6.84}	& 7.32	& 2.50	& 0.47	& 2.21	& 0.27	& 1.50	& 10.00 \\
& SFCFusion \cite{chen2024sfcfusion}   &6.40	&7.08	&2.21		&0.62	& 2.23  &0.37	&1.36  &11.29 \\
& MMDRFuse \cite{deng2024mmdrfuse}     & 6.63 	& 8.30 	& 2.63 	& 0.51 	& 2.91 	& 0.13 	& 1.05 	& 11.00 \\
& UMF-CMGR \cite{wang2022unsupervised}    & 6.55 	& 8.84 	& 2.44 	& 0.63 	& 2.22 	& 0.41 	& 1.47 	& 9.86 \\
&FISCNet\cite{zheng2024frequency}	  &6.72	&13.61	&3.97		&\underline{0.77}  &\textbf{2.94}	&\underline{0.64}	&1.34 &\underline{3.71} \\
&RPFNet\cite{guan2025residual} 	 &\textbf{7.38}	&\underline{13.63}	&\underline{4.06}		&0.74  &2.03	&0.55	&1.63 &4.57 \\
& ISFM(Ours)    & 6.76 & \textbf{13.65} & \textbf{4.10} & \textbf{0.80} & 2.61 & \textbf{0.66} & \textbf{1.68} & \textbf{1.71} \\ \midrule
\multirow{7}{*}
& ReCoNet \cite{huang2022reconet}   & 7.07	 & 9.31	& 3.90		& 0.53  & 2.15  &0.38	& 1.57  & 8.57  \\
& DeFusion \cite{liang2022fusion}    & 6.93	& 8.99	& 3.59		& 0.50	 & 2.06  & 0.37	& 1.35  & 11.86 \\
& U2Fusion \cite{xu2020u2fusion}   &6.78 	&12.38 	&5.00 	 &0.52   &1.82 	&0.49 	&1.27   &9.43 \\
& TarDAL \cite{liu2022target}   & \underline{7.24}	& 10.65  & 4.22		& 0.54  & 2.13  	& 0.40	& 1.47  & 7.00 \\
& IRFS \cite{wang2023interactively}   & 7.01	& 10.79	& 4.08		& 0.57 & 2.01	& 0.46	& \underline{1.58}  & 7.43 \\
& AdaFuse \cite{gu2023adafuse}      &7.09	&7.64	&3.44		&0.41	&1.52  &0.25	&1.33  &13.43\\
{\textbf{RoadScene}} & DATFuse \cite{tang2023datfuse}   & 6.77	& 11.81	 & 4.22		&\underline{0.58} &\textbf{2.49}	&0.47	&1.16  &7.43  \\
& LRRNet \cite{li2023lrrnet}    &  7.14	& 12.79   & 4.80	&0.49 &1.95	&0.35	&1.57  & 8.14 \\
& DDFM \cite{zhao2023ddfm}   & 6.56	& 7.71	& 2.94		& 0.54 & \underline{2.20}	& 0.32	& 1.12  & 12.00 \\
& FusionMamba \cite{xie2024fusionmamba}     & 7.20	 & 13.85   & \underline{5.41}	& 0.53  & 2.10		& 0.40	& 1.53 & 5.57 \\
& SFCFusion \cite{chen2024sfcfusion}  &6.87 	&9.65 	&3.76 	 	&0.52 &1.92	&0.37 	&1.35 &11.86 \\
& MMDRFuse  \cite{deng2024mmdrfuse}    &  7.09	 & 10.38	& 4.26		& 0.49   & 2.15	& 0.14	& 1.04  & 11.00 \\
& UMF-CMGR \cite{wang2022unsupervised}   & 7.06 	& 10.82 & 4.18	& 0.57  & 2.13 	& 0.48 	& 1.48   &7.00\\
&FISCNet\cite{zheng2024frequency}	&7.09	&\underline{14.62}	&5.31		&0.57  &2.20	&\underline{0.52}	&1.26  &\underline{4.43} \\
&RPFNet	\cite{guan2025residual}  &\textbf{7.40}	&13.83	&5.05		&0.53  &1.94	&0.46	&1.54  &5.86 \\
& ISFM(Ours)         & 7.11  & \textbf{14.87} & \textbf{5.50} &\textbf{0.59} & 2.16  & \textbf{0.53} & \textbf{1.59} & \textbf{2.00} \\ \bottomrule
\end{tabular}
\end{adjustbox}
\label{tab:performance_ivif}
\vspace{-4mm}
\end{table*}
\begin{table*}[htbp]
 \renewcommand{\arraystretch}{0.91}
\centering
\caption{Quantitative results of the MIF task. The best performance is marked in \textbf{bold} and the second-best performance is \underline{underlined}.}
\begin{adjustbox}{width=0.8\textwidth}
\begin{tabular}{@{}c|l|ccccccccc@{}}
\toprule
\textbf{Dataset} & \textbf{Method} & \textbf{EN(↑)} & \textbf{SF(↑)} & \textbf{AG(↑)} & \textbf{VIF(↑)} & \textbf{MI(↑)} & \(\mathbf{Q^{AB/F}}\)(↑) & \textbf{SCD(↑)} & \textbf{Avg.R(↓)} \\ \midrule
\multirow{7}{*}
& ReCoNet \cite{huang2022reconet}   &4.36	&16.5	&5.70	&0.42	&1.51	&0.30	&1.14	&13.29   \\
& DeFusion \cite{liang2022fusion}   &5.58	&29.17	  &9.13	&0.58	&2.04	&0.62	&0.95	&6.43 \\
& U2Fusion \cite{xu2020u2fusion}    &5.37	&19.86	&6.66		&0.46	&1.80 &0.41	&0.52 & 12.14 \\
& TarDAL \cite{liu2022target}   &5.72	&27.54	&9.11	&0.31	&1.56	&0.30	&0.77	&10.86  \\
& IRFS \cite{wang2023interactively}   &5.60	&23.80	&7.46	&0.62	&2.27	&0.54	&1.47	&6.29 \\
& AdaFuse \cite{gu2023adafuse}    &6.18	&19.91	&6.60	&\underline{0.68}	&2.23	&0.47	&1.18	&7.29   \\
{\textbf{MRI-PET}}& DATFuse \cite{tang2023datfuse}   &6.00 	&\textbf{45.91}	 &\underline{11.19}	 &0.42	&1.93	&0.54	&0.84	&6.57 \\
& LRRNet \cite{li2023lrrnet}   &5.57	&15.88	&5.32	&0.33	&1.65	&0.18	&0.25	&14.71 \\
& DDFM \cite{zhao2023ddfm}    &5.26	&21.10	&6.93	&0.68	&\underline{2.35}	&0.55	&1.28	&7.00 \\
& FusionMamba \cite{xie2024fusionmamba}    &4.95	&\underline{42.05}	&\textbf{12.20}	&0.59	&2.02	&\underline{0.65}	&\underline{1.62}	&\underline{4.86}\\
& SFCFusion \cite{chen2024sfcfusion}   &5.37	&40.47	&7.89	&0.43	&2.33	&0.36	&0.75 & 9.14 \\
& MMDRFuse \cite{deng2024mmdrfuse}     &5.56	&33.03	&8.14	&0.50	&2.01	&0.51	&1.06	&8.14 \\
& UMF-CMGR \cite{wang2022unsupervised}    &6.20 	&30.72	&8.44	&0.41	&1.72	&0.43	&0.35	&9.86 \\
&FISCNet\cite{zheng2024frequency}	&\underline{6.22}	&23.72	&9.04		&0.57 &1.81	&0.59	&1.10  &6.71  \\
&RPFNet\cite{guan2025residual}	&\textbf{7.04}	&17.97	&5.59		&0.52 &1.83	&0.51	&1.04  &9.29  \\
& ISFM(Ours)      &5.61	&33.13	&10.30	&\textbf{0.73}	&\textbf{2.40}	&\textbf{0.70}	&\textbf{1.66}	&\textbf{2.57} \\ \midrule
\multirow{7}{*}
& ReCoNet \cite{huang2022reconet}   &4.42	&21.45	&5.71	&0.42	&1.91	&0.40	&1.11	&9.14  \\
& DeFusion \cite{liang2022fusion}   &4.67	    &19.76	 &5.09	&0.42	&1.94	&0.37	&1.07	&9.57 \\
& U2Fusion \cite{xu2020u2fusion}   &4.69	&20.05	&5.74	&0.34	&1.66	&0.40	&0.55 &11.29  \\
& TarDAL \cite{liu2022target}   &5.79	&22.72	&7.45	&0.23	&1.49	&0.28	&0.44	&11.14  \\
& IRFS \cite{wang2023interactively}   &5.15	&26.93	  &6.53	&0.41	&1.97	&0.52	&1.45	&5.86 \\
& AdaFuse \cite{gu2023adafuse}     &5.42	&18.92	 &4.74	&0.41	&\textbf{2.26}	&0.33	&0.94	&9.71 \\
{\textbf{MRI-CT}} & DATFuse \cite{tang2023datfuse}   &5.05	  &27.61	&6.29	&0.46	&2.09	&0.48	&1.20	&5.86 \\
& LRRNet \cite{li2023lrrnet}    &4.90	  &20.20	&4.80	&0.37	&1.86	&0.33	&0.18	&12.29 \\
& DDFM \cite{zhao2023ddfm}   &4.31	&19.64	  &4.81	  &0.40	  &\underline{2.18}	  &0.37	&1.03	&10.57 \\
& FusionMamba \cite{xie2024fusionmamba}      &4.40	&34.80 &\textbf{7.67}	&0.50	&1.99	&0.52	&\underline{1.57}	&4.86 \\
& SFCFusion \cite{chen2024sfcfusion}   &4.49	&\textbf{37.86}	&6.48	&0.29	&2.16	&0.34	&0.70  &9.14 \\
& MMDRFuse \cite{deng2024mmdrfuse}     &4.58	&\underline{35.11}  	&6.02	&0.39	&2.01	&0.37	&1.25	&7.57 \\
& UMF-CMGR \cite{wang2022unsupervised}   & 5.56	    &34.27	 &7.40   	&0.32	&1.78	&0.38	&0.39	&9.29 \\
&FISCNet\cite{zheng2024frequency}	&\underline{6.56}	&33.49 &8.54		&\underline{0.50}  &1.90	&\underline{0.53}	&1.28 &  \textbf{3.86}\\
&RPFNet\cite{guan2025residual} 	&\textbf{6.78}	&18.10	&4.57		&0.37 &1.80	&0.45	&1.04  & 10.29  \\
& ISFM(Ours)         &4.61	&33.10	&\underline{7.51}	&\textbf{0.53}	&2.02	&\textbf{0.54}	&\textbf{1.72}	&\underline{4.14} \\ \midrule
\multirow{7}{*}
& ReCoNet \cite{huang2022reconet} &3.71	&10.61	    &3.40	&0.35	&1.55	&0.28	&0.53	&15.14  \\
& DeFusion \cite{liang2022fusion}    &4.86	    &14.81	 &4.55	&0.57	&2.10	&0.54	&0.65	&9.43  \\
& U2Fusion \cite{xu2020u2fusion}   &4.61	&13.53	&4.22	&0.44	&1.84	&0.44	&0.90  &12.14 \\
& TarDAL \cite{liu2022target}   &5.59	&16.67	 &5.63	&0.32	&1.55	&0.23	&0.88	&10.29  \\
& IRFS \cite{wang2023interactively}   & 5.39	&14.49	    &4.63	&0.55	&2.11	&0.53	&1.34	&7.57 \\
& AdaFuse \cite{gu2023adafuse}      & 5.42	&14.22	    &4.61	&0.56	&2.17	&0.55	&1.47	&6.86 \\
{\textbf{MRI-SPECT}} & DATFuse \cite{tang2023datfuse}   &5.23	  &23.40	  &6.27	  &0.51	&\underline{2.24}	&0.60	&0.88	&5.43  \\
& LRRNet \cite{li2023lrrnet}   &5.12	&11.62	&3.78	&0.36	&1.80	&0.20	&0.52	&14.00  \\
& DDFM \cite{zhao2023ddfm}   &4.79	&15.20	&4.77	&0.55	&2.22	&0.58	&1.53	&6.57  \\
& FusionMamba \cite{xie2024fusionmamba}    &4.79	&\underline{24.89}	    &\textbf{7.18}	&\underline{0.64}	&2.16	&\underline{0.68}	&\underline{1.53}	&\underline{3.86} \\
& SFCFusion \cite{chen2024sfcfusion}   &4.70	&\textbf{27.37}	&4.54	&0.42	&2.28	&0.31	&0.50  &10.00 \\
& MMDRFuse \cite{deng2024mmdrfuse}    & 5.15	    &21.85	 &4.88	&0.49	&2.08	&0.49	&1.18	&7.71 \\
& UMF-CMGR \cite{wang2022unsupervised}   &5.52	     &22.92	  &5.25	&0.38	&1.81	&0.44	&0.78	&8.86\\
&FISCNet\cite{zheng2024frequency}	  &\underline{6.21}	&17.18	&6.00		&0.53  &1.96	&0.61	&1.16  &5.71  \\
&RPFNet	\cite{guan2025residual}   &\textbf{6.96}	&15.06	 &4.57		&0.45  &1.81	&0.59	&0.96  &8.14  \\
& ISFM(Ours)        &5.02	&20.74	 &\underline{6.44}	&\textbf{0.65}	&\textbf{2.28}	&\textbf{0.70}	&\textbf{1.75}	&\textbf{3.14} \\ \bottomrule
\end{tabular}
\end{adjustbox}
\label{tab:performance_mif}
\vspace{-4mm}
\end{table*}
\subsection{Experimental Setups}
\label{sec:setup}
\subsubsection{Datasets and Metrics}
We conduct extensive experiments to evaluate our method on six widely-used MMIF datasets.
For IVIF, we adapt three dataset: MSRS \cite{tang2022piafusion}, FMB \cite{liu2023multi} and RoadScene \cite{xu2020u2fusion}.
We train and test our framework on MSRS dataset which consists of 1083 pairs for training and 361 pairs for testing.
We also test on the FMB and RoadScene datasets which consists of 280 pairs and 221 pairs for testing.
These datasets include a diverse range of images captured in both the day and night, including various objects such as people, cars and bikes.
For MIF, we adopt Harvard Medical dataset\footnote{http://www.med.harvard. edu/AANLIB/home.html}.
The Harvard Medical dataset is a public medical image website, from which we selected 150 pairs for testing, including 50 pairs of MRI-CT images, 50 pairs of MRI-PET images and 50 pairs of MRI-SPECT images.
Note that, fine-tuning is not applied to the FMB, RoadScene and Harvard Medical dataset to fully verify the generalization performance.

We use eight metrics to measure the fusion results: Entropy (EN), Spatial Frequency (SF), Average Gradient (AG), Visual Information Fidelity (VIF), Mutual Information (MI), Edge-based similarity measure (\(Q^{AB/F}\)), Sum of the Correlations of Differences (SCD) and Average Rank (Avg.Rank).
The details of these metrics can be found in \cite{ma2019infrared,park2023cross}.
\subsubsection{Implementation Details}
In this work, all the experiments are performed with the PyTorch toolbox and one GeForce RTX 3090 GPU.
The training samples are randomly cropped into 128$\times$128 patches in the preprocessing stage.
The number of epochs for training is set to 100 and the batch size is set to 4.
To optimize our model, we adopt the Adam \cite{kingma2014adam} optimizer with the initial learning rate set to $10^{-4}$.
The learning rate follows a warmup strategy with a cosine decay.
For the hyper-parameters, the number of VSSM and K are set to 2 and 4, respectively.
The dimension of channels in VSSM is set to 128.
The expansion factor \(\eta\) is set to 2.
\(\alpha\), \(\lambda\) and \(\gamma\) are set to 5, 5 and 0.5, respectively.
\subsection{Comparisons with Other Methods}
\label{sec:IVIF}
In this subsection, we compare our method with 15 \sArt methods, including ReCoNet \cite{huang2022reconet}, DeFusion \cite{liang2022fusion}, U2Fusion \cite{xu2020u2fusion}, TarDAL \cite{liu2022target}, IRFS \cite{wang2023interactively}, AdaFuse\cite{gu2023adafuse}, DATFuse \cite{tang2023datfuse}, LRRNet \cite{li2023lrrnet}, DDFM \cite{zhao2023ddfm}, FusionMamba \cite{xie2024fusionmamba}, SFCFusion\cite{chen2024sfcfusion}, MMDRFuse\cite{deng2024mmdrfuse}, UMF-CMGR \cite{wang2022unsupervised}, FISCNet \cite{zheng2024frequency} and RPFNet \cite{guan2025residual}.
\subsubsection{Qualitative Analysis}
For IVIF, as shown in \figref{fig:visual_msrs}, \figref{fig:visual_fmb} and \figref{fig:visual_road}, we compare the fusion results of several methods on the MSRS, FMB and RoadScene dataset, respectively.
It can be observed that our ISFM effectively preserves thermal radiation information from the infrared image, while maintaining texture details of the targets.
For instance, in \figref{fig:visual_msrs}, the foreground target, such as the person in the red box, is clearly highlighted with sharp details.
However, the fused images obtained by other methods (such as U2Fusion, TarDAL, LRRNet, {\em etc.}) are visually unsatisfactory.
For example, in the green box of \figref{fig:visual_fmb}, the license plate details are poorly retained.
The fused images produced by our method exhibit rich texture details and a superior visual quality.

For MIF, we show the qualitative comparison in \figref{fig:visual_mripet}, \figref{fig:visual_mrict} and \figref{fig:visual_mrispect}.
It can be observed that U2Fusion and MMDRFuse methods fail to preserve the functional information from the PET, CT and SPECT images well.
LRRNet and DDFM methods suffer from an obvious color distortion.
The proposed method can better preserve both structural details and functional information from the source images, generating fused images with a high visual quality.
\subsubsection{Quantitative Analysis}
For IVIF, the quantitative results are shown in \tabref{tab:performance_ivif}.
The proposed method achieves the best Avg.Rank on three datasets, demonstrating that our method outperforms other methods.
Compared with the methods that rely solely on spatial fusion, such as ReCoNet, DeFusion and U2Fusion, our ISFM shows clear advantages in all metrics.
Compared with other spatial-frequency fusion methods, such as SFCFusion, FISCNet and RPFNet, our method improves the overall performance and ranks the first in Avg.Rank.

For MIF, the quantitative results are presented in \tabref{tab:performance_mif}.
Note that our ISFM is not fine-tuned on the MIF dataset while methods like U2Fusion, SFCFusion and FusionMamba are trained on it.
Although some metrics like EN and SF do not yield the best results, our ISFM achieves leading performance in terms of Avg.Rank.
The experimental results show that our method can achieve robust performance.
It further demonstrates that our method has a strong generalization ability across different datasets.
\begin{figure}[t]
\centering
  \includegraphics[width=0.95\linewidth]{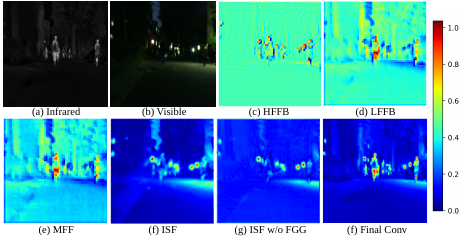}
  \caption{Visualization of feature maps obtained from different modules. (a) Infrared image; (b) Visible image; (c) HFFB; (d) LFFB; (e) MFF; (f) ISF; (e) ISF without FGG; (f) Final convolutional layer.}
  \label{fig:visual_map}
\end{figure}
\subsubsection{Visualization of Spatial and Frequency Fusion Features}
\label{sec:visual}
To verify the effectiveness of the proposed modules, we further visualize the extracted features of different modules. The visualizations are shown in \figref{fig:visual_map}.
It can be observed that low-frequency features predominantly capture the overall image structure, whereas high-frequency features emphasize edge details.
The proposed MFF effectively integrates these frequency domain characteristics.
Next, we display the feature maps of the proposed ISF.
It can be seen that the features highlight prominent target details while preserving essential background information.
In addition, without the frequency domain interaction provided by the FGG, the features fail to enhance texture details, resulting in a less comprehensive representation.
Finally, the features of the final convolutional layer preserve the distinct characteristics of both spatial and frequency domains, leading to a significant improvement in the overall quality of the fused image.
\begin{table}
\centering
\caption{Ablation study of the key modules on MSRS dataset.}
\begin{adjustbox}{width=0.9\linewidth}
\begin{tabular}{@{}lccccccc@{}}
\toprule
&Configurations               & EN   & SF  & AG & SCD & VIF  & $Q^{AB/F}$ \\ \midrule
I  & Baseline               & 6.46 	& 10.69 	& 3.39 	& 1.68 	& 0.87 	& 0.59 \\
II  & Baseline+MFF               & 6.63 	& 11.24	& 3.7 	& 1.70 	& 0.99 	& 0.66 \\
III & Baseline+FGM                        & 6.63 	& 11.17 	& 3.58 	& 1.73 	& 0.95 	& 0.65       \\   	
IV & Baseline+ISF            & 6.65  	& 11.25 	& 3.66	& 1.74 	& 0.96 	& 0.66 \\
V  & Baseline+MFF+FGM            & 6.67 	& 11.32 	& 3.69 	& 1.77 	& 1.00 	& 0.68   \\ \midrule 	 	 	 	
&Ours                       & \textbf{6.70} & \textbf{11.42} & \textbf{3.77} & \textbf{1.79} &\textbf{1.01}  &\textbf{0.68}\\ \bottomrule
\end{tabular}
\end{adjustbox}
\label{tab:abs_key}
\end{table}
\begin{table}
\centering
\caption{Analysis of FGM on MSRS dataset.}
\begin{adjustbox}{width=0.9\linewidth}
\begin{tabular}{@{}lccccccc@{}}
\toprule
&Configurations               & EN   & SF  & AG & SCD & VIF  & $Q^{AB/F}$ \\ \midrule
I  & Addition                                  & 6.65 	& 11.11 	& 3.55 	& 1.68 	& 0.89 	& 0.65  \\ 					
II &  Concatenation                         & 6.66 	& 11.13 	& 3.56 	& 1.68 	& 0.91 	& 0.65     \\					
III  &   Cross-attention                  & 6.65 	& 11.22 	& 3.68 	& 1.69 	&0.98	& 0.66 \\			
IV & FGM              & \textbf{6.70} & \textbf{11.42} & \textbf{3.77} & \textbf{1.79} &\textbf{1.01}  &\textbf{0.68}    \\ \bottomrule
\end{tabular}
\end{adjustbox}
\label{tab:abs_FGCM}
\end{table}
\begin{table}
\centering
\caption{Analysis of FGG on MSRS dataset.}
\begin{adjustbox}{width=0.9\linewidth}
\begin{tabular}{@{}lccccccc@{}}
\toprule
&Configurations               & EN   & SF  & AG & SCD & VIF  & $Q^{AB/F}$ \\ \midrule
I  & W/o interaction                                  & 6.67 	& 11.32	& 3.69 	& 1.77 	& 1.00  	& 0.68  \\	
II &  Only frequency                           & 6.67 	& 11.34	& 3.70 	& 1.76 	& 1.01  	& 0.68    \\				
III & FGG             & \textbf{6.70} & \textbf{11.42} & \textbf{3.77} & \textbf{1.79} &\textbf{1.01}  &\textbf{0.68}    \\ \bottomrule
\end{tabular}
\end{adjustbox}
\label{tab:abs_FGGM}
\end{table}
\begin{table}[t]
\centering
\caption{Ablation study of frequency domain fusion on MSRS dataset.}
\begin{adjustbox}{width=0.9\linewidth}
\begin{tabular}{@{}c|ccc|cccccc@{}}
\toprule
 & LFFB & HFFB  & FGG  & EN   & SF  & AG & SCD & VIF  & $Q^{AB/F}$  \\ \midrule
 I &  & &    & 6.63 	& 11.17 	& 3.58 	& 1.73 	& 0.95 	& 0.65     \\
II & \checkmark &  &  &  6.66 	& 11.13 	& 3.55 	& 1.76 	& 0.99 	& 0.67\\
III &  &    \checkmark        &  & 6.64 	& 11.23 	& 3.57 	& 1.70 	& 0.95 	& 0.65\\
IV & \checkmark    & \checkmark &  & 6.67 	& 11.32 	& 3.69 	& 1.77 	& 1.00 	& 0.68\\
V & \checkmark & \checkmark & \checkmark        & \textbf{6.70} & \textbf{11.42} & \textbf{3.77} & \textbf{1.79} &\textbf{1.01}  &\textbf{0.68}\\
 \bottomrule
\end{tabular}
\end{adjustbox}
\label{tab:abs_ff}
\end{table}
\subsection{Ablation Studies}
\label{sec:ablation}
We conduct a series of ablation experiments on the MSRS dataset to evaluate the effectiveness of the proposed method.
\subsubsection{Effect of Key Modules}
To validate the effectiveness of the proposed modules, we conduct ablation experiments under different configurations.
As shown in \tabref{tab:abs_key}, the baseline is the model without additional modules.
We concatenate the modality-specific features to generate the fused images.
In the configuration II, we incorporate MFF into the baseline, and the results show a significant improvement.
In the configuration III, we introduce FGM to the baseline, which enhances the fusion results compared with the baseline.
The configuration IV builds upon the configuration III by adding FGG, with all metrics showing notable improvements.
The results indicate that the interaction between the spatial and frequency domains effectively improves the quality of the fusion.
In the configuration V, we combine MFF with FGM, but without the spatial-frequency interaction.
Compared with the final model, the lack of interaction leads to a suboptimal fusion.
These experimental results demonstrate the effectiveness of the proposed modules.
\begin{figure}[t]
\centering
\includegraphics[width=0.9\linewidth]{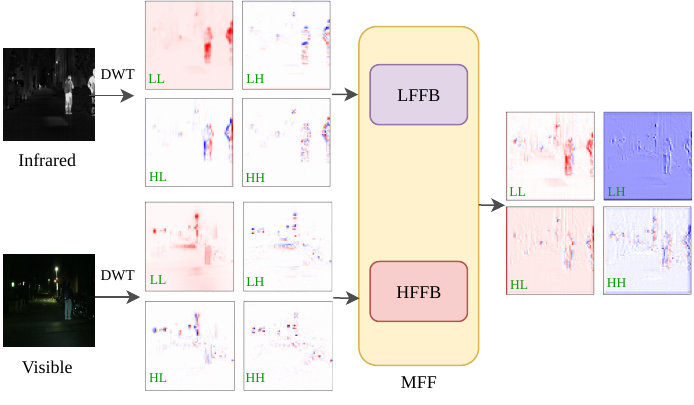}
\caption{Visualization of DWT decomposition (LL, LH, HL, HH) of visible and infrared inputs and the fused features produced by our MFF.}
\label{fig:visual_dwt}
\end{figure}
\begin{figure}[t]
\centering
\includegraphics[width=1.0\linewidth]{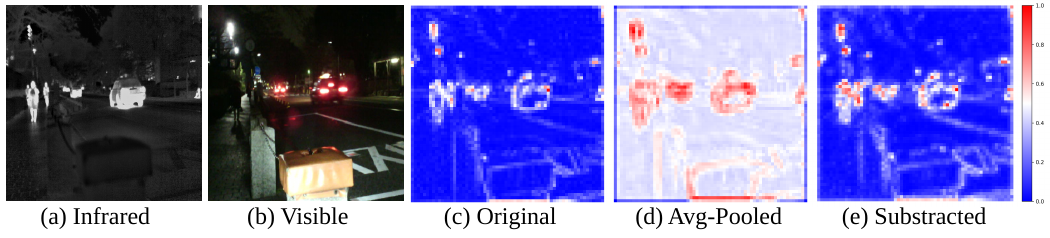}
\caption{Visualization of the proposed high-frequency enhancement operation.}
\label{fig:visual_hf}
\end{figure}
\begin{figure}[t]
\centering
\includegraphics[width=0.86\linewidth]{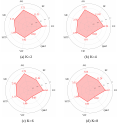}
\caption{Ablation study with different $K$ on MSRS dataset.}
\label{fig:abs_num}
\end{figure}
\subsubsection{Analysis of Interactive Spatial-Frequency Fusion}
In the proposed ISF, FGM plays a crucial role in extracting complementary information in the spatial domain.
To assess the impact of this module, we remove FGM and instead perform an element-wise addition for extracting complementary information, as shown in the configuration I of \tabref{tab:abs_FGCM}.
In the configuration II, we replace FGM with a channel-wise concatenation followed by a fully connected layer.
In the configuration III, we utilize cross-attention for the same purpose.
The results show that the proposed FGM significantly outperforms other methods.
These findings underscore the effectiveness of FGM in enhancing complementary information.

In addition, we investigate the impact of FGG, as shown in \tabref{tab:abs_FGGM}.
The first configuration does not incorporate any interaction between the spatial and frequency domains.
In the configuration II, we introduce frequency information into the interaction process.
But the gating weights are only driven from frequency features.
The result shows that it still falls short of achieving optimal interaction.
Finally, in the configuration III, we employ the proposed FGG.
It achieves the best performance.
These results show that ISF enables a more robust integration between the spatial and frequency domains.
\subsubsection{Analysis of Frequency Domain Fusion}
We further conduct ablation experiments to verify the effectiveness of the proposed frequency domain fusion, as shown in \tabref{tab:abs_ff}.
In Exp. I, without incorporating frequency domain information, the fusion method produces the worst results.
In Exp. II, low-frequency information is fused using LFFB, while high-frequency features are directly summed.
In Exp. III, HFFB is utilized for high-frequency fusion with low-frequency features summed.
In Exp. IV, we apply FFB to integrate frequency domain information.
Finally, we perform the spatial-frequency interaction through FGG.
By integrating all modules, our method achieves optimal performance.
These results show the effectiveness of our method in frequency domain fusion.

To visually validate the effectiveness of our frequency-domain fusion mechanism, we conduct two kinds of visualization experiments.
First, we show the DWT decomposition of the source images and the corresponding features fused by the proposed MFF.
As shown in \figref{fig:visual_dwt}, the LL, LH, HL and HH sub-bands reveal the complementary nature of different modalities.
Our MFF effectively integrates these characteristics, producing fused features with rich structures and high-frequency details.
Second, we visualize the effect of the high-frequency enhancement operation,
Conceptually, this operation is analogous to the unsharp masking in signal processing.
In fact, it enhances the high-frequency components by subtracting the average-pooled (blurred) features from the original signals.
As shown in \figref{fig:visual_hf}, the visualization confirms this effect.
It shows stronger activations along object boundaries and suppressed responses in smooth regions.
\subsubsection{Effect of the Hyper-parameter $K$}
To investigate the impact of the hyper-parameter $K$, we conduct experiments on MSRS dataset.
As illustrated in \figref{fig:abs_num}, the overall performance significantly improves as $K$ increases up to 4.
Beyond this, the performance improves on SF, AG, MI but decreases on other metrics.
With further increases, most metrics exhibit significant declines.
To balance the performance and computational efficiency, we set $K=4$ as default.
\begin{table}
\renewcommand{\arraystretch}{0.9}
\setlength{\tabcolsep}{4pt}
\centering
\small
\caption{Parameter count of the proposed framework.}
\begin{adjustbox}{width=0.4\linewidth}
\begin{tabular}{@{}lcc@{}}
\toprule
Module & Params (M) \\
\midrule
MSE & 0.9752 \\
MFF & 4.4186 \\
ISF & 3.1635 \\
Other Layers & 0.5905 \\
\midrule
Total & 9.148 \\
\bottomrule
\end{tabular}
\end{adjustbox}
\label{tab:params_model}
\end{table}
\begin{table}
\renewcommand{\arraystretch}{0.85}
\setlength{\tabcolsep}{4pt}      
\centering
\caption{Comparison of computational efficiency on MSRS dataset.}
\begin{adjustbox}{width=0.85\linewidth}
\begin{tabular}{@{}llccc@{}}
\toprule
Method & Time (s) & GFLOPs (G) & Params (M) \\
\midrule
IRFS \cite{wang2023interactively}            & 1.45   & 74  & 0.242   \\
FusionMamba \cite{xie2024fusionmamba}  & 0.17   & 125 & 164.539 \\
DeFusion \cite{liang2022fusion}      & 0.33   & 143 & 7.874   \\
UMF-CMGR \cite{wang2022unsupervised}  & 0.03   & 193 & 0.629   \\
DDFM \cite{zhao2023ddfm}         & 63.46  & 139k  & 552.814 \\
AdaFuse \cite{gu2023adafuse}    & 1.65   & 431 & 538.529 \\
SFCFusion \cite{chen2024sfcfusion}  & 0.21   & 68  & 0.074   \\
RPFNet \cite{zheng2024frequency}            & 0.03   & 47  & 0.072   \\
FISCNet  \cite{guan2025residual}            & 0.07   & 101 & 0.324   \\
Ours       & 0.28  & 371    & 9.148 \\	
\bottomrule
\end{tabular}
\end{adjustbox}
\label{tab:comp_effi}
\end{table}
\subsection{Computational Efficiency Analysis}
\label{sec:comp_effi}
In this subsection, we conduct a comprehensive computational efficiency analysis.
\tabref{tab:params_model} shows the parameter distribution across the key modules.
In addition, we compare our method with several state-of-the-art methods in terms of inference time, parameters, and GFLOPs.
As shown in \tabref{tab:comp_effi}, our method maintains a moderate level of GFLOPs and parameters, achieving a favorable balance between the model complexity and performance.
\subsection{Evaluation of Downstream Tasks}
\label{sec:downstream}
In this subsection, we further evaluate the effectiveness of our method in two downstream tasks, \emph{i.e.}, object detection and semantic segmentation.
\subsubsection{Object Detection Evaluation}
\label{sec:detection}
\begin{figure*}[htbp]
\centering
\includegraphics[width=1.0\linewidth]{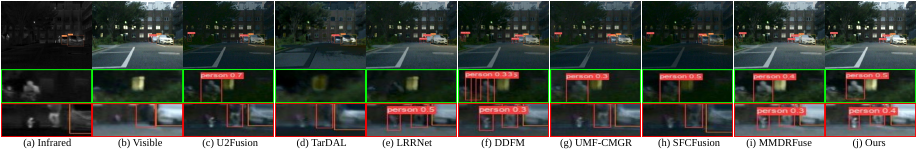}
\caption{Comparison of object detection results on MSRS dataset. The second row shows the enlarged regions in the first row.}
\label{fig:detect}
\end{figure*}

We first perform object detection on MSRS dataset.
YOLOv5 \cite{redmon2016yolo} is employed to evaluate the detection performance.
\figref{fig:detect} illustrates the advantages of our ISFM in improving object detection performance.
In the ``00002D'' image, the detector fails to identify pedestrians from the fusion results of other methods due to the complexity of the scene and the degradation of illumination.
In contrast, our ISFM effectively integrates the complementary information of infrared and visible images, ensuring all pedestrians are successfully detected.
We further use quantitative metrics to measure the object detection performance: precision, recall, mean average precision (mAP).
The results shown in \tabref{tab:detect} demonstrate that the proposed ISFM can improve object detection performance, thereby facilitating computer vision tasks under adverse environmental conditions.

\subsubsection{Semantic Segmentation Evaluation}
\label{sec:segmentation}
\begin{figure*}[htbp]
\centering
\includegraphics[width=1.0\linewidth]{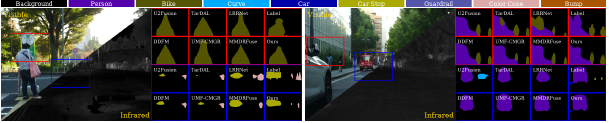}
\caption{Comparison of semantic segmentation results on MSRS dataset.}
\label{fig:seg}
\end{figure*}

We also conduct the semantic segmentation on MSRS dataset.
It includes annotations for nine object categories (\emph{i.e.}, background, bump, color cone, guardrail, curve, bike, person, car stop and car).
We retrain DeeplabV3+ \cite{chen2018encoder} on infrared images, visible images, and the fused images from each method to evaluate their performances.
As shown in \figref{fig:seg}, our method produces segmentation maps with a better region completeness.
We further report the mean intersection-over-union (mIoU) in \tabref{tab:seg}.
The results demonstrate that our method can better preserve edge information from the source modalities, leading to the improved segmentation accuracy.
\begin{table}
\centering
\caption{Object detection performance on MSRS dataset.}
\begin{adjustbox}{width=0.8\linewidth}
\begin{tabular}{@{}lcccc@{}}
\toprule
&Precision   & Recall   & mAP@0.50   \\ \midrule
ReCoNet \cite{huang2022reconet} & 0.972	&0.853	&0.935 	 	 \\
DeFusion \cite{liang2022fusion} &0.982	&0.947	&0.976 	      \\
U2Fusion \cite{xu2020u2fusion}  &0.962	&0.961	&0.980 	 	  \\
TarDAL \cite{liu2022target} &0.931	&0.835	&0.922 		      \\
IRFS \cite{wang2023interactively} & 0.981	&\underline{0.961}	&0.978 	 	      \\
AdaFuse \cite{gu2023adafuse}  &0.966	&0.923	&0.976 	 	 \\
DATFuse \cite{tang2023datfuse}  &0.977 	&0.934 	&0.969  	 	 \\
LRRNet \cite{li2023lrrnet} &0.980 	&0.949 	&0.983  	      \\
DDFM \cite{zhao2023ddfm}  &0.978	&0.956	&0.981	 	 \\
FusionMamba \cite{xie2024fusionmamba}  &0.956 	&0.928 	&0.966  	 	 \\
SFCFusion \cite{chen2024sfcfusion} &0.962	&0.928	&0.977	 	      \\		
MMDRFuse \cite{deng2024mmdrfuse} &\underline{0.982}	&0.946	&\underline{0.983}	 	      \\
UMF-CMGR \cite{wang2022unsupervised}  &0.964	&0.876	&0.953 	 	 \\
FISCNet\cite{zheng2024frequency}	&0.967	&0.965	&0.971 \\
RPFNet	\cite{guan2025residual}	&0.973	&0.958	&0.976 \\
Ours & \textbf{0.984}	&\textbf{0.962}	&\textbf{0.985}   \\ \bottomrule
\end{tabular}
\end{adjustbox}
\label{tab:detect}
\end{table}
\begin{table}
\renewcommand{\arraystretch}{0.95}
\setlength{\tabcolsep}{2pt}      
\centering
\caption{Semantic segmentation performance on MSRS dataset.}
\begin{adjustbox}{width=0.9\linewidth}
\begin{tabular}{@{}lccccccccc|c@{}}
\toprule
     & Unl & Car & Per & Bik & Cur & CS & GD & CC & Bu & mIoU \\
    \midrule
    Visible & 96.92 & 79.89 & 34.53 & 61.98 & 18.99 & 50.48 & 41.47 & 47.78 & 60.73 & 54.75 \\
    Infrared & 96.57 & 73.58 & 46.99 & 52.63 & 29.94 & 24.51 & 52.81 & 22.98 & 47.79 & 49.76 \\
    \midrule
    ReN\cite{huang2022reconet}  & 97.33 & 81.48 & 57.48 & 59.09 & 36.11 & 49.48 & 55.48 & 36.13 & 57.77 & 58.93 \\
    DeF\cite{liang2022fusion} & 97.79 & 86.21 & 64.91 & 63.90 & 28.33 & 64.58 & 58.39 & 46.51 & 60.67 & 63.48 \\
    U2F\cite{xu2020u2fusion}  & 97.84 & 86.20 & 65.51 & 65.74 & 33.68 & 61.29 & 35.29 & 49.49 & 61.33 & 61.82 \\
    TarD\cite{liu2022target} & 97.67 & 84.70 & 63.10 & 61.96 & 28.41 & 60.56 & 38.98 & 48.56 & 57.62 & 60.17 \\
     IRFS \cite{wang2023interactively}  & 97.69 & 85.41 & 59.43 & \underline{67.39} & 42.45 & 53.71 & 63.82 & 52.68 & 65.78 & 65.37 \\
     AdaF\cite{gu2023adafuse}  & 97.30 & 83.39 & 57.18 & 63.11 & 34.60 & 47.25 & 55.24 & 49.13 & 44.04 & 59.03 \\
     DAT \cite{tang2023datfuse} & 97.64 & 84.07 & 60.51 & 63.62 & 40.15 & 56.21 & 61.05 & 48.22 & 61.18 & 63.63 \\
      LRR \cite{li2023lrrnet}  & 97.51 & 82.65 & 63.16 & 65.22 & 36.46 & 52.53 & 52.71 & 49.48 & 59.57 & 62.14 \\
    DDFM \cite{zhao2023ddfm} & 97.64 & 84.84 & 60.31 & 65.09 & 41.37 & 52.00 & 62.52 & 51.13 & 63.67 & 64.29 \\
    FuMa\cite{xie2024fusionmamba} & 98.02 & \underline{87.93}& \underline{65.88} & 66.17 & 43.58 & 65.45 & 58.01 & 52.75 & 68.13 & 67.32 \\
    SFC \cite{chen2024sfcfusion}  & 97.84 & 86.20 & 65.51 & 65.74 & 33.68 & 61.29 & 35.29 & 49.49 & 61.33 & 61.82\\
    MDR \cite{deng2024mmdrfuse}  & \underline{98.05} & 87.91 & 65.68 & 67.20 & \underline{46.12} & \underline{68.01} & 65.05 & \underline{53.66} & \underline{69.51} &\underline{69.02} \\
    UMF \cite{wang2022unsupervised} & 97.79 & 86.21 & 64.91 & 63.90 & 28.33 & 64.58 & 58.39 & 46.51 & 60.67 & 63.48\\
    FISC\cite{zheng2024frequency} & 97.67 & 84.29 & 59.89 & 64.55 & 41.80 & 56.98 & 66.17 & 49.05 & 63.97 & 64.93\\
    RPF\cite{guan2025residual} & 97.15 & 82.23 & 28.43 & 64.93 & 36.06 & 55.41 & \textbf{66.86} & 49.19 & 64.18 & 60.49 \\
    \midrule
    Ours & \textbf{98.15} & \textbf{88.66} & \textbf{66.30} & \textbf{68.69} & \textbf{49.88} & \textbf{69.00} & \underline{66.22} & \textbf{54.29} & \textbf{73.60} & \textbf{70.53} \\
    \bottomrule
\end{tabular}
\end{adjustbox}
\label{tab:seg}
\end{table}
\section{Conclusion}
In this paper, we propose a novel interactive spatial-frequency fusion Mamba framework for multi-modal image fusion.
More specially, we propose a Multi-scale Frequency Fusion (MFF) to integrate frequency domain information on multiple scales.
To exploit the complementarity of domain-specific characteristics, we propose an Interactive Spatial-Frequency Fusion (ISF).
It incorporates frequency features to guide spatial fusion across modalities and leverages Mamba to enhance complementary representations of spatial-frequency domains.
Extensive experiments on six MMIF datasets validate the effectiveness of the proposed method.

\bibliographystyle{IEEEtran}
\bibliography{IEEEabrv,refs}

\vfill

\end{document}